\documentclass[final, 12pt]{colt2018} % Anonymized submission
\usepackage{times}
\usepackage{epstopdf}
\newcommand{\norm}[1]{\left\lVert #1 \right\rVert}
\newcommand{\cost}{\mathrm{cost}}
\newcommand{\sumt}{\sum_{t=1}^T}

\usepackage{etoolbox}

\ifcsmacro{theorem}{}{
\newtheorem{theorem}{Theorem}
\newtheorem{corollary}[theorem]{Corollary}
\newtheorem{lemma}[theorem]{Lemma}
\newtheorem{proposition}[theorem]{Proposition}

\newtheorem{definition}[definition]{Definition}
\newtheorem{algorithm}[algorithm]{Algorithm}

}

\newtheorem{innercustomgeneric}{\customgenericname}
\providecommand{\customgenericname}{}
\newcommand{\newcustomtheorem}[2]{%
  \newenvironment{#1}[1]
  {%
   \renewcommand\customgenericname{#2}%
   \renewcommand\theinnercustomgeneric{##1}%
   \innercustomgeneric
  }
  {\endinnercustomgeneric}
}

\newcustomtheorem{customthm}{Theorem}
\newcustomtheorem{customlemma}{Lemma}

\usepackage{algorithm}
\usepackage{algorithmic}
\usepackage{color}
\DeclareMathOperator*{\argmin}{arg\!min}

\newcommand{\ouralg}{Online Balanced Descent}
\newcommand{\ourack}{OBD}

\usepackage{comment}
\usepackage{ifthen}
\newboolean{showcomments}
\setboolean{showcomments}{true}
\newcommand{\niangjun}[1]{  \ifthenelse{\boolean{showcomments}}
{ \textcolor{red}{(Niangjun:  #1)}} {}  }
\newcommand{\adam}[1]{  \ifthenelse{\boolean{showcomments}}
{ \textcolor{red}{(Adam:  #1)}} {}  }
\newcommand{\gautam}[1]{  \ifthenelse{\boolean{showcomments}}
{ \textcolor{red}{(Gautam:  #1)}} {}  }
\newcommand{\addcites}[0]{  \ifthenelse{\boolean{showcomments}}
{ \textcolor{red}{(Add citation(s))}} {}  }
\newcommand{\addcite}[0]{  \ifthenelse{\boolean{showcomments}}
{ \textcolor{red}{(Add citation(s))}} {}  }

\graphicspath{{figures/}}
  \coltauthor{\Name{Niangjun Chen} $^a$\thanks{Niangjun Chen and Gautam Goel contributed equally to this work.}
  \Email{chennj@ihpc.a-star.edu.sg}\\
  \Name{Gautam Goel} $^b$\footnotemark[1] \Email{ggoel@caltech.edu}\\
  \Name{Adam Wierman} $^b$\Email{adamw@caltech.edu}\\
  \addr{$^a$ Institute of High Performance Computing} \mbox{}\hfill \addr{$^b$ California Institute of Technology}
  }

\title{ Smoothed Online Convex Optimization in High Dimensions  \quad \\ via Online Balanced Descent}
\begin{document}
\maketitle

\begin{abstract}
We study \emph{smoothed online convex optimization}, a version of online convex optimization where the learner incurs a penalty for changing her actions between rounds. Given a $\Omega(\sqrt{d})$ lower bound on the competitive ratio of any online algorithm, where $d$ is the dimension of the action space, we ask under what conditions this bound can be beaten. We introduce a novel algorithmic framework for this problem, \ouralg\ (\ourack), which works by iteratively projecting the previous point onto a carefully chosen level set of the current cost function so as to balance the switching costs and hitting costs. We demonstrate the generality of the \ourack\ framework by showing how, with different choices of ``balance,'' \ourack\ can improve upon state-of-the-art performance guarantees for both competitive ratio and regret; in particular,  \ourack\ is the first algorithm to achieve a dimension-free competitive ratio, $3 + O(1/\alpha)$,  for locally polyhedral costs, where $\alpha$ measures the ``steepness'' of the costs.  We also prove bounds on the dynamic regret of OBD when the balance is performed in the dual space that are dimension-free and imply that \ourack\ has sublinear static regret.

\end{abstract}
%\section{}
%\subsection{}

\section{Introduction}
\label{sec: introduction}
%!TEX root = LookingGlass.tex

In this paper we develop a new algorithmic framework, \ouralg\ (\ourack), for online convex optimization problems with switching costs, a class of problems termed smoothed online convex optimization (SOCO).  Specifically, we consider a setting where a learner plays a series of rounds $1,2,\ldots,T$. In each round, the learner observes a convex cost function $f_t$, picks a point $x_t$ from a convex set $\mathcal{X}$, and then incurs a \emph{hitting cost} $f_t(x_t)$. Additionally, she incurs a \emph{switching cost} for changing her actions between successive rounds, $\|x_t - x_{t-1}\|$, where $\|\cdot\|$ is a norm.  

This setting generalizes classical Online Convex Optimization (OCO), and has received considerable attention in recent years as a result of the recognition that switching costs play a crucial role in many learning, algorithms, control, and networking problems.  In particular, many applications have, in reality, some cost associated with a change of action that motivates the learner to adopt ``smooth'' sequences of actions.  For example, switching costs have received considerable attention in the $k$-armed bandit setting  \citep{agrawal1990multi, guha2009multi, koren2017multi} and the core of the Metrical Task Systems (MTS) literature is determining how to manage switching costs, e.g., the $k$-server problem \citep{borodin1992, borodin2005}. %More generally, even when there is no measurable cost to switching, if there is concept drift in a penalized estimation problem, then it is natural to use a regularization term (switching cost) to control the drift of the estimator.  

Outside of learning, SOCO has received considerable attention in the networking and control communities.  In these problems there is typically a measurable cost to changing an action.  For example, one of the initial applications where SOCO was adopted is the dynamic management of service capacity in data centers \citep{lin2011, lu2013simple}, where the wear-and-tear costs of switching servers into and out of deep power-saving states is considerable.  Other applications where SOCO has seen real-world deployment are the dynamic management of routing between data centers \citep{lin2012,wang2014exploring}, management of electrical vehicle charging \citep{kim2014real}, video streaming \citep{joseph2012jointly}, speech animation \citep{kim2015}, multi-timescale control \citep{goel2017thinking}, power generation planning \citep{badiei2015online}, and the thermal management of System-on-Chip (SoC) circuits \citep{zanini2009multicore,zanini2010online}.

\paragraph{High-dimensional SOCO.} An important aspect of nearly all the problems mentioned above is that they are \emph{high-dimensional}, i.e., the dimension $d$ of the action space is large.  For example, in the case of dynamic management of data centers the dimension grows with the heterogeneity of the storage and compute nodes in the cluster, as well as the heterogeneity of the incoming workloads. However, the design of algorithms for high-dimensional SOCO problems has proven challenging, with fundamental lower bounds blocking progress.  

Initial results on SOCO focused on finding competitive algorithms in the low-dimensional settings. Specifically, \cite{lin2011} introduced the problem in the one-dimensional case and gave a 3-competitive algorithm.  A few years later, \cite{bansal2015} gave a 2-competitive algorithm, still for the one-dimensional case.  Following these papers, \cite{antoniadis2016} claimed that SOCO is equivalent to the classical problem of Convex Body Chasing \cite{Friedman1993}, in the sense that a competitive algorithm for one problem implies the existence of a competitive problem for the other. Using this connection, they claimed to show the existence of a constant competitive algorithm for two-dimensional SOCO. However, their analysis turned out to have a bug and their claims have been retracted \citep{errata}. 
However, the connection to Convex Body Chasing does highlight a fundamental limitation.  In particular, it is not possible to design a competitive algorithm for high-dimensional SOCO without making restrictions on the cost functions considered since, as we observe in Section \ref{sec: model}, for general convex cost functions and $\ell_2$ switching costs, the competitive ratio of any algorithm is $\Omega(\sqrt{d})$. 

The importance of high-dimensional SOCO problems in practical applications has motivated the ``beyond worst-case'' analysis for SOCO as a way of overcoming the challenge of designing constant competitive algorithms in high dimensions.  To this end, \cite{lin2012, andrew2013, chen2015, badiei2015online, chen2016} all explored the value of predictions in SOCO, highlighting that it is possible to provide constant-competitive algorithms for high-dimensional SOCO problems using algorithms that have predictions of future cost functions, e.g., \cite{lin2012} gave an algorithm based on receding horizon control that is $1+O(1/w)$-competitive when given $w$-step lookahead. Recently, this was revisited in the case quadratic switching costs by \cite{li2018online}, which gives an algorithm that combines receding horizon control with gradient descent to achieve a competitive ratio that decays exponentially in $w$. 

In addition to the stream of work focused on competitive ratio, there is a separate stream of work focusing on the development of algorithms with small regret.  With respect to classical \emph{static} regret, where the comparison is with the  fixed, static offline optimal, \cite{andrew2013} showed that SOCO is no more challenging than OCO.  In fact, many OCO algorithms, e.g., Online Gradient Descent (OGD), obtain bounds on regret of the same asymptotic order for SOCO as for OCO.  However, the task of bounding \emph{dynamic} regret in SOCO is more challenging and, to this point, the only positive results for dynamic regret rely on the use of predictions.  %In particular, \cite{li2018online} gives a receding horizon gradient descent algorithm that has dynamic regret that shrinks exponentially in the size of the lookahead $w$.  But, we are not aware of any positive results for dynamic regret that do not use predictions. \niangjun{commented out as we already talked about this work in the previous paragraph.}

There have been a number of attempts to connect the communities focusing on regret and competitive ratio over the years.  \cite{Blum2000} initiated this direction by providing an analytic framework that connects OCO algorithms with MTS algorithms, allowing the derivation of regret bounds for MTS algorithms in the OCO framework and competitive analysis of OCO algorithms in the MTS framework.  Two breakthroughs in this direction occurred recently.  First, \cite{buchbinder2012unified} used a primal-dual technique to develop an algorithm that, for the first time, provided a unified approach for algorithm design across competitive ratio and regret in the MTS setting, over a discrete action space. Second, a series of recent papers \citep{abernethy2010regularization, buchbinder2014competitive, bubeck2017} used techniques based on Online Mirror Descent (OMD) to provide significant advances in the analysis of the $k$-server and MTS problems. However, again there is a fundamental limit to these unifying approaches in the setting of SOCO. \cite{andrew2013} shows that no individual algorithm can simultaneously be constant competitive and have sublinear regret.  Currently, the only unifying frameworks for SOCO rely on the use of predictions, and use approaches based on receding horizon control, e.g., \cite{chen2015,badiei2015online, chen2016,li2018online}. 

\paragraph{Contributions of this paper.} The prior discussion highlights the challenges associated with designing algorithms for high-dimensional SOCO problems,
%In particular, the development of algorithms for high-dimensional SOCO remains an open and active topic, 
both in terms of competitive ratio and dynamic regret.  In this paper, we introduce a new, general algorithmic framework, \ouralg\ (\ourack), that yields (i) an algorithm with a dimension-free, constant competitive ratio for locally polyhedral cost functions and $\ell_2$ switching costs and (ii) an algorithm with a dimension-free, sublinear static regret that does not depend on the size of the gradients of the cost functions. In both cases, \ourack\ achieves these results without relying on predictions of future cost functions -- the first algorithm to achieve such a bound on competitive ratio outside of the one-dimensional setting.

The key idea behind \ourack\ is %that, while many online optimization algorithms descend toward the minimizer of the cost function in each round, it is \niangjun{commented out as OGD and other online alg doesn't really move towards minimum at each step, only the 1-d memoryless alg. in Bansal2015.}
%to focus on the geometry of the level sets to guide the choice of an action in each round.  This insight leads \ourack\ 
to move using a projection onto a carefully chosen level set at each step chosen to ``balance'' the switching and hitting costs incurred.  The 
%projection is performed using a mirror map, as in Online Mirror Descent (OMD), and the \niangjun{commented out since mirror projection is quite common in the learning field.}
form of ``balance'' is chosen carefully based on the performance metric.  In particular, our results use a ``balance'' of the switching and hitting costs in the primal space to obtain results for the competitive ratio and a ``balance''  of the switching costs and the gradient of the hitting costs in the dual space to obtain results for dynamic regret. The resulting \ourack\ algorithms are efficient to implement, even in high dimensions. They are also \emph{memoryless}, i.e., do not use any information about previous cost functions. 

The technical results of the paper bound the competitive ratio and dynamic regret of \ourack.  In both cases we obtain results that improve the state-of-the-art. In the case of \emph{competitive ratio}, we obtain the first results that break through the $\sqrt{d}$ barrier without the use of predictions.  In particular, we show that \ourack\ with $\ell_2$ switching costs yields a constant, dimension-free competitive ratio for locally polyhedral cost functions, i.e. functions which grow at least linearly away from their minimizer.  Specifically, in Theorem \ref{threeCR} we show that \ourack\ has a competitive ratio of $3+O(1/\alpha)$, where $\alpha$ bounds the ``steepness'' of the costs.  Note that \cite{bansal2015} shows that no memoryless algorithm can achieve a competitive ratio better than $3$ for locally polyhedral functions. 
%Our approach extends to more general classes of cost functions and switching costs as well; we give the first competitive algorithm when the cost functions and switching costs are arbitrary norms 
By equivalence of norms in finite dimensional space, our algorithm is also competitive when the switching costs are arbitrary norms (though the exact competitive ratio may depend on $d$).  Our proof of this result depends crucially on the geometry of the level sets. 
%The fact that the cost functions are locally polyhedral gives that the level sets are not too narrow, and the key geometric insight in our analysis is that this roundness implies bounds on the deviation from the optimal offline costs at each step. Lemma  \ref{lem: potential-change} is the key technical result capturing this insight.  \niangjun{commented out as it is too early for reader to understand this detail of the proof.}

%Our proof technique uses a potential function argument; we break the per-round analysis into two cases depending on whether our algorithm or the offline optimal is closer to the minimizer, and bound the change in potential in each case. The first case is fairly simple; due to the fact that our algorithm balances the hit cost and movement cost, it is straightforward to show that the per-step cost we incur is at most a constant times the cost incurred by the offline optimal. The second case is more involved. The key geometric insight in our analysis is captured by Lemma  \ref{lem: potential-change}, where we show that in the second case the change in potential is actually negative. This offsets the hit cost and movement cost incurred by our online algorithm, so the total per-step cost we incur is actually non-positive, allowing us to sidestep a direct comparison with the offline cost.

In the case of \emph{dynamic regret}, we obtain the first results that provide sub-linear dynamic regret without the use of predictions.  Further, the bounds on dynamic regret we prove are independent of the size of the gradient of the cost function. %, something that current state-of-the-art results in OCO do not achieve.  \niangjun{commented out to avoid triggering the hardcore learning theory people into pointing out that we cheat by having one-step advantage.}
Specifically, in Corollary \ref{cor: online-projection-regret} we show that \ourack\ has dynamic regret bounded by $O(\sqrt{LT})$, where $L$ is the total distance traveled by the offline optimal. When comparing to a static optimal, \ourack\ achieves sublinear regret of $O(\sqrt{DT})$ where $D$ is the diameter of the feasible set. The proof again makes use of a geometric interpretation of \ourack\ in terms of the level sets.  In particular, the projection onto the level sets allows \ourack\ to be ``one step ahead'' of Online Mirror Descent. Further, \ourack\ carefully choses the step sizes to balance the switching cost and marginal hitting cost in the dual space.

\section{Online Optimization with Switching Costs}
\label{sec: model}
%!TEX root = LookingGlass.tex

We study online convex optimization problems with switching costs, a class of problems often termed \emph{Smoothed Online Convex Optimization (SOCO)}. An instance of SOCO consists of a fixed convex decision/action space $\mathcal{X} \subset \mathbb{R}^d$, a norm $\|\cdot \|$ on $\mathbb{R}^d$, and a sequence of non-negative convex cost functions $f_t$, where $t = 1 \ldots T$ and $f_t(x) = +\infty$ for all $x \notin \mathcal{X}$. 

At each time $t$, an online learner observes the cost function $f_t$ and chooses a point $x_t$ in $\mathcal{X}$, incurring a \emph{hitting cost} $f_t(x_t)$ and a \emph{switching cost} $\|x_t - x_{t-1}\|$. The total cost incurred is thus 
\begin{equation}
\cost(ALG)=\sumt f_t(x_t) + \norm{x_t - x_{t-1}},
\end{equation}
where $x_t$ are the decisions of online algorithm $ALG$.  While we state this as unconstrained, constraints are incorporated via the decision space $\mathcal{X}$. Note that the decision variable could be taken to be matrix-valued instead of vector valued; similarly the switching cost could be any matrix norm.

Importantly, we have allowed the learner to see the current cost function when deciding on an action, i.e., the online algorithm observes $f_t$ before picking $x_t$.  This is standard when studying switching costs due to the added complexity created by the coupling of the actions $x_t$ due to the switching cost term, e.g., \cite{andrew2013,bansal2015}, but it is different than the standard assumption in the OCO literature. While allowing the algorithm to see $f_t$ in the standard OCO setting would make the problem trivial, in the SOCO setting considerable complexity remains -- the choice in round $t$ of the offline optimal depends on $f_s$ for all $s>t$.  Giving the algorithm complete information about $f_t$ isolates the difficulty of the problem, eliminating the performance loss that comes from not knowing $f_t$ and focusing on the performance loss due to the coupling created by the switching costs. This is natural since it is easy to bound the extra cost incurred due to lack of knowledge of $f_t$.  In particular, as done in \cite{Blum2000} and \cite{buchbinder2012unified}, the penalty due to not knowing $f_t$ is 
$$ f_t(x_t) - f_{t}(x_{t+1}) \leq \nabla \langle f_t(x_t), x_t - x_{t+1}\rangle \leq \norm{\nabla f_t(x_t)}_2 \norm{x_t - x_{t+1}}_2.$$
Since cost functions are typically assumed to have a bounded gradient in the the OCO literature, this equation gives a translation of the results in this paper to the setting where $f_t$ is not known.  

Note that the SOCO model is closely related to the Metrical Task System (MTS) literature.  In MTS, the cost functions $f_t$ can be arbitrary, the feasible set $\mathcal{X}$ is discrete, and the movement cost can be any metric distance. Due to the generality of the MTS setting, the results are typically pessimistic. For example, \cite{borodin1992} shows that the competitive ratio of any deterministic algorithm must be $\Omega(n)$ where $n$ is the size of $\mathcal{X}$, and \cite{blum1992} shows an $\Omega(\sqrt{\log(n)/\log\log(n)})$ lower bound for randomized algorithms. In comparison, SOCO restricts both the cost functions and the feasible space to be convex, though the decision space is continuous rather than discrete.

\paragraph{Performance Metrics.}
The performance of online learning algorithms in this setting is evaluated by comparing the cost of the algorithm to the cost achievable by the \textit{offline optimal}, which makes decisions with advance knowledge of the cost functions. The cost incurred by the offline optimal is  
\begin{equation}
\cost(OPT) = \min_{x \in \mathcal{X}^T} \sumt f_t(x_t) + \norm{x_t - x_{t-1}}.
\label{eqn: offline-cost}
\end{equation}

Sometimes, it is desirable to constrain the power of the offline optimal when performing comparisons.  One common approach for this, which we adopt in this paper, is to constrain the movement the offline optimal is allowed \citep{blum1992,buchbinder2012unified,Blum2000}. This is natural for our setting, given the switching costs incurred by the learner. Specifically, define the $L$-constrained offline optimal as the offline optimal solution with switching cost upper bounded by $L$, i.e., the minimizer of the following (offline) problem:
\begin{align*}
OPT(L) = \min_{x \in \mathcal{X}^T}\quad \sumt f_t(x_t) + \norm{x_t - x_{t-1}} \qquad
 \text{subject to} \quad \sumt \norm{x_t - x_{t-1}} \le L.  
\end{align*}
For large enough $L$, $OPT(L)=OPT$.  Specifically, $L = \sumt \norm{x_t^* - x_{t-1}^*}$ guarantees that $\cost(OPT(L)) = \cost(OPT)$. Further, for small enough $L$, $OPT(L)$ corresponds to the \emph{static optimal}, $OPT^{STA}$, i.e., the optimal static choice such that $x_t=x_1$. Since the movement cost of the static optimal (assuming it takes one step from $x_0$) is $\norm{x^{STA}- x_0} \le D$, $\cost(OPT(D)) \le \cost(OPT^{STA})$. Therefore $\cost(OPT(L))$ interpolates between the cost of the dynamic offline optimal to the static offline optimal as $L$ varies.  

When comparing an online algorithm to the (constrained) offline optimal cost, two different approaches are common.  The first, most typically used in the online algorithms community, is to use a multiplicative comparison. This yields the following definition of the \emph{competitive ratio}.

\begin{definition}
An online algorithm $ALG$ is \textbf{$C$-competitive} if, for all sequences of cost functions $f_1, \ldots, f_T$, we have $\cost(ALG) \le C \cdot \cost(OPT). $
\end{definition}

As discussed in the introduction, there has been considerable work focused on designing algorithms that have a competitive $C$ that is constant with respect to the dimension of the decision space, $d$.  In contrast to the multiplicative comparison with the offline optimal cost in the competitive ratio, an additive comparison is most common in the online learning community.  This yields the following definition of \emph{dynamic regret}.

\begin{definition}
The \textbf{$L$-(constrained) dynamic regret} of an online algorithm $ALG$ is $\rho_L(T)$ if for all sequences of cost functions $f_t, \ldots, f_T$, we have 
$\cost(ALG) - \cost(OPT(L)) \le \rho_L(T).$ $ALG$ is said to be \textbf{no-regret} against OPT(L) if $\rho_L(T)$ is sublinear.
\end{definition}

As discussed above, $OPT(L)$ interpolates between the offline optimal $OPT$ and the offline static optimal $OPT^{STA}$.  There are many algorithms that are known to achieve $O(\sqrt{T})$ static regret, the best possible given general convex cost functions, e.g., online gradient descent \cite{zinkevich2003} and follow the regularized leader \cite{xiao2010}. While these results were proven initially for OCO, \cite{andrew2013} shows that the same bounds hold for SOCO. In contrast, prior work on dynamic regret has focused primarily on OCO, e.g., \cite{herbster2001, bianchi2012,hall2013dynamical}.  For SOCO, the only positive results for SOCO consider algorithms that have access to predictions of future workloads, e.g., \cite{lin2012, chen2015, chen2016}.

For both competitive ratio and dynamic regret, it is natural to ask what performance guarantees are achievable.  The following lower bounds (proven in the appendix) follow from connections to Convex Body Chasing first observed by \cite{antoniadis2016}. 

\begin{proposition} \label{prop: lowerbound}
The competitive ratio of any online algorithm for SOCO is $\Omega(\sqrt{d})$ with $\ell_2$ switching costs and $\Omega(d)$ with $\ell_\infty$ switching costs. The dynamic regret is $\Omega(d)$ in both settings.
\end{proposition}

\section{\ouralg\ (\ourack)}
\label{sec: alg-meta}

The core of this paper is a new algorithmic framework for online convex optimization that we term \emph{\ouralg} (\ourack). \ourack\ represents the first algorithmic framework that applies to both competitive analysis and regret analysis in SOCO problems and, as such, parallels recent results that have begun to provide unified algorithmic techniques for competitiveness and regret in other areas.  For example,  \cite{buchbinder2012unified} and \cite{Blum2000} provided frameworks that bridged competitiveness and regret in the context of MTS problems, which have a discrete action spaces, and \cite{blum2002static} did the same for decision making problems on trees and lists. 

\ourack\ builds both on ideas from online algorithms, specifically the work of \cite{bansal2015}, and online learning, specifically Online Mirror Descent (OMD) \citep{nemirovskii1983problem, warmuth1997continuous, bubeck2015}.  We begin this section with the geometric intuition underlying the design of \ourack\ in Section \ref{sec:alg-geometry}.  Then, in Section \ref{sec:alg-description}, we describe the details of the algorithm. Finally, in Section \ref{sec:alg-examples} we provide illustration of how the choice of the mirror map impacts the behavior of the algorithm. 

\subsection{A Geometric Interpretation}
\label{sec:alg-geometry}

%A classical approach for online optimization is to move greedily towards the minimizer $v_t$ during each round \niangjun{OGD doesn't move towards $v_t$ as well}.  Assuming this structure, the only design choice left is how to pick the step-size at each round. In contrast, 
The key insight that drives the design of \ourack\ is that it is the geometry of the level sets, not the location of the minimizer, which should guide the choice of $x_t$. %To see this, consider a cost function whose level sets are elongated ellipses. Suppose the previous point $x_{t-1}$ is located near the end of such an ellipse. Moving directly towards the minimizer incurs an unnecessarily high switching cost; a much cheaper action would be to be \emph{lazy} and simply project onto the level set. This incurs the same hitting cost, but much less switching cost. 
This reasoning leads naturally to an algorithm for SOCO that, at each round, projects the previous point $x_{t-1}$ onto a level of the current cost function $f_t$. However, the question of ``which level set?'' remains.  This choice is where the name \ouralg\ comes from -- \ourack\ chooses a level set that \emph{balances} the switching cost and hitting costs, where the notion of balance used is the heart of the algorithm.  

To make the intuition described above more concrete, assume for the moment that the switching cost is defined by the $\ell_2$ norm. Suppose the algorithm has decided to project onto the $l$-level set of $f_t$ (we show how to pick $l$ in the next section). Then the action of the algorithm is the solution of the following optimization problem: 
\begin{align*}
\text{minimize} \quad \frac{1}{2}\norm{x - x_{t-1}}^2 \qquad \text{subject to } \quad f_t(x) \le l.
\end{align*}
Now, let $\eta_t$ be the optimal dual variable for the inequality constraint. By the first order optimality condition, $x_t$ needs to satisfy 
$$x_t = x_{t-1} - \eta_t \nabla f_t(x_t).$$
This resembles a step of Online Gradient Descent (OGD), except that the update direction is the gradient $\nabla f_t(x_t)$ instead of $\nabla f_{t-1}(x_{t-1})$, and the step size $\eta_t$ is allowed to vary. Hence in this setting \ourack\ can be seen as a ``one step ahead'' version of OGD with time-varying step sizes.

\begin{figure}[t]
\floatconts
  {fig:subfigex}
  {\caption{\textit{An illustration of the difference between \ourack\ and OMD assuming the mirror map $\Phi(x) = \frac{1}{2}\norm{x}^2_2$. While OMD steps in a direction normal to the contour line of $f_{t-1}(\cdot)$ at $x_{t-1}$, \ourack\  steps in a direction normal to the contour line of $f_t(\cdot)$ at $x_t$. }}
  \label{fig: comparison}}
  {%
    \subfigure[\textit{A step taken by OMD. Contour lines represent the sub-level sets of $f_{t-1}$.}]{\label{fig: omd_step}%
      \hspace{.5in} \includegraphics[width=0.23\linewidth]{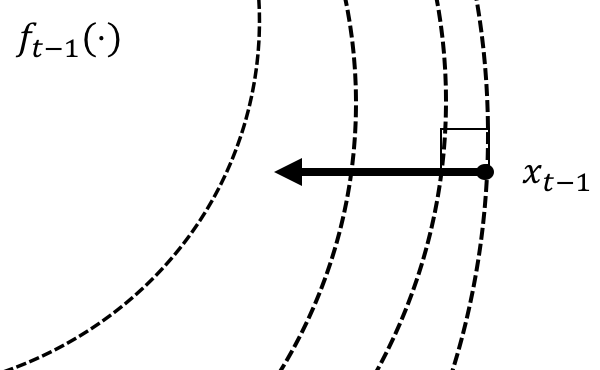}\hspace{.5in}}%
    \qquad \quad
    \subfigure[\textit{A step taken by \ourack. Contour lines represent the sub-level sets of $f_{t}$.}]{\label{fig: ouralg_step}%
      \hspace{.5in}\includegraphics[width=0.23\linewidth]{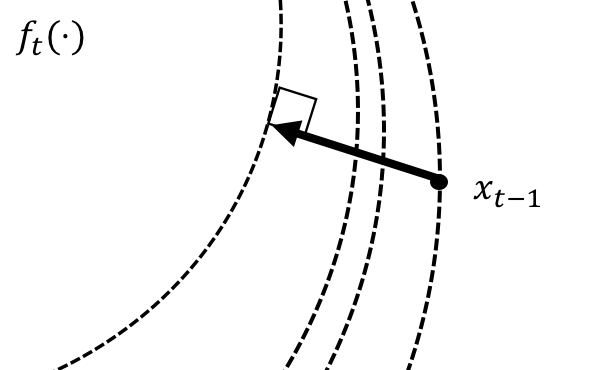}\hspace{.5in}}
 \vspace{-.2in}
 }
 \vspace{-.15in}
\end{figure}

For more general switching costs a similar geometric intuition can be obtained using a mirror map $\Phi$ with respect to the norm $\| \cdot \|$.  Here, $x_t$ is the solution of the following optimization in dual space where, given a convex function $\Phi$, $D_\Phi(x, y)$ is the Bregman divergence between $x$ and $y$, i.e., $D_\Phi(x, y) = \Phi(x) - \Phi(y) - \nabla \Phi(y)^T(x-y)$: 
\begin{align*}
\text{minimize}  \quad D_{\Phi}(x, x_{t-1}) \qquad
\text{subject to} \quad f_t(x) \le l.
\end{align*}
As before, let $\eta_t$ be the optimal dual variable for the inequality constraint. The first order optimality condition implies that $x_t$ must satisfy 
\begin{equation}
\nabla \Phi(x_t) = \nabla \Phi(x_{t-1}) - \eta_t \nabla f_t (x_t).
\label{eqn: mirror-descent-step}
\end{equation}
The form of \eqref{eqn: mirror-descent-step} is similar to a ``one step ahead'' version of OMD with time varying $\eta_t$, i.e., the update direction in the dual space is in the gradient $\nabla f_t(x_t)$ instead of the gradient $\nabla f_{t-1}(x_{t-1})$. The implicit form of the update has been widely used in online learning, e.g., \cite{kivinen1997exponentiated, kulis2010implicit}.

Figure \ref{fig: comparison} illustrates the difference between \ourack\ and OMD when $\Phi(x) = \frac{1}{2}\norm{x}^2_2$: \ourack\ is normal to the destination whereas OMD is normal to the starting point. Intuitively, this is why the guarantees we obtain for \ourack\ are stronger than what previous descent-based approaches have obtained in this setting -- it is better to move in the direction determined by the level set where you land, than the direction determined by the level set where you start. 

\subsection{A Meta Algorithm}
\label{sec:alg-description}

The previous section gives intuition about one key aspect of the algorithm, the projection onto a level-set.  But, in the discussion above we assume we are projecting onto a specific $l$-sublevel set. The core of \ourack\ is that this sublevel set is determined endogenously in order to ``balance'' the switching and hitting costs, as opposed to a fixed exogenous schedule of step-sizes like is typical in many online descent algorithms. Informally, the operation of \ourack\ is summarized in the \emph{meta algorithm} in Algorithm \ref{alg: online-projection}, which uses the operator $\Pi^\Phi_K(x) : \mathbb{R}^n \rightarrow K$ to denote the Bregman projection of $x$ onto a convex set $K$, i.e., $\Pi^\Phi_K(x) = \argmin_{y \in K} D_\Phi(y, x)$, where $\Phi$ is $m$-strongly convex and $M$-Lipschitz smooth in $\norm{\cdot}$, i.e., $ \frac{m}{2} \norm{x-y}^2 \le D_\Phi(x,y) \le \frac{M}{2} \norm{x-y}^2.$

We term Algorithm \ref{alg: online-projection} a meta algorithm because the general framework given in Algorithm \ref{alg: online-projection} can be instantiated with different forms of ``balance'' in order to perform well for different metrics.  More specifically, the notion of ``balance'' in the Step \ref{alg:meta-balance} that is appropriate varies depending on whether the goal is to perform well for competitive ratio or for regret.  

Our results in this paper highlight two different approaches for defining \emph{balance} in \ourack\ based on either balancing the switching cost with the hitting cost in either the primal or dual space.  We  balance costs in the primal space to yield a constant, dimension-free competitive algorithm for locally polyhedral cost functions (Section \ref{sec: alg-cr}), and balance in the dual space to yield a no-regret algorithm (Section \ref{sec: alg-regret}). We summarize these two approaches in the following and then give more complete descriptions in the corresponding technical sections.  

%\adam{Need to reincorporate the following: Let $x(l) = \Pi^\Phi_{K_l}(x_{t-1})$.  }

\begin{algorithm}[t]
	\begin{algorithmic}[1]
			\REQUIRE Starting point $x_0$, mirror map $\Phi$.
			\FOR{$t=1, \ldots, T$}
			\STATE Choose a sublevel set $K_l = \{ x \mid f_t(x) \le l\}$ to ``balance'' the switching and hitting costs.  \label{alg:meta-balance}
			  \STATE Set $x_t = \Pi^{\Phi}_{K_l}(x_{t-1})$.
              \label{alg: gradient-step}
			\ENDFOR
	\end{algorithmic}
\caption{\ouralg\ (\ourack), Meta Algorithm}
\label{alg: online-projection}
\end{algorithm}

\begin{itemize}
\item{\textbf{Primal \ouralg}}.  The algorithm we consider in Section \ref{sec: alg-cr} instantiates Algorithm \ref{alg: online-projection} by choosing $l$ such that $x(l) = \Pi^\Phi_{K_l}(x_{t-1})$ achieves balance between the switching cost with the hitting cost in the \emph{primal} space. Specifically, for some fixed $\beta>0$, choose $l$ such that either $x(l) = \argmin_x f_t(x)$ and $\norm{x(l) - x_{t-1}} < \beta l $, or the following is satisfied:
\begin{equation}
\norm{x(l) - x_{t-1}} = \beta l
\label{eqn: primal-balance}
\end{equation}

\item{\textbf{Dual \ouralg}}. The algorithm we consider in Section \ref{sec: alg-regret} instantiates Algorithm \ref{alg: online-projection} by balancing the switching cost with the size of the gradient in the \emph{dual} space.  Specifically, for some fixed $\eta$, we choose $l$ such that 
\begin{equation}
\norm{\nabla \Phi(x(l)) - \nabla \Phi(x_{t-1})}_* = \eta \norm{\nabla f_t(x(l))}_*,
\label{eqn: dual-balance}
\end{equation}
\end{itemize}

The final piece of the algorithm is computational.  Note that the algorithm is \emph{memoryless}, i.e., it does not use any information about previous cost functions.  Thus, the only question about efficiency is whether the appropriate $l$ can be found efficiently.  The following lemmas verify that, indeed, it is possible to compute $l$, and thus implement \ourack, efficiently.  

\begin{lemma}
		The function $g(l) = \norm{x(l) - x_{t-1}}$ is continuous in $l$. 
		\label{lem: continuity-distance}
\end{lemma}

\begin{lemma}
Consider $\Phi$ and $f_t$ that are continuously differentiable on $\mathcal{X}$. The function \\ $h(l) = \frac{\norm{\nabla \Phi(x_{t-1}) - \nabla \Phi(x(l)) }_*}{\norm{\nabla{f_t(x(l))}}_*}$ is continuous in $l$. 
\label{lem: continuity-ratio}
\end{lemma}

The continuity of $g(l)$ and $h(l)$ in $l$ is enough to guarantee efficient implementation of Primal and Dual \ourack\ because it shows that an $l$ satisfying the balance conditions in the algorithms exists and, further, can be found to arbitrary precision via bisection.  Proofs are included in the appendix.

\subsection{Examples}
\label{sec:alg-examples}

An important part of the design of an \ourack\ algorithm is the choice of the mirror map.  
%Recall that \ourack\ needs to choose a mirror map that is both strongly convex and Lipschitz smooth for the norm defined by the switching cost. Clearly, 
Different choices of a mirror map $\Phi$ can lead to very different behavior by the resulting algorithms. To highlight this, and give intuition for the impact of the choice, we describe three examples of mirror maps below.  These examples focus on mirror maps that are commonly used for OMD, and they highlight interesting connections between the \ourack\ framework and classical online optimization algorithms like OGD \citep{zinkevich2003} and Multiplicative Weights \citep{arora2012}.

\paragraph{Euclidean norm squared:} Consider $\Phi(x) = \frac{1}{2}\norm{x}_2^2$, which is both 1-strongly convex and 1-Lipschitz smooth for the $\ell_2$ norm. Note that $\nabla \Phi(x) = x$.  Then, the first order condition \eqref{eqn: mirror-descent-step} is 
\begin{equation}
x_t = x_{t-1} - \eta_t \nabla f_t(x_t).
\label{eqn: one-step-ahead-GD}
\end{equation}
Interestingly, this can be interpreted as a ``one-step ahead'' OGD (illustrated in Figure \ref{fig: comparison}).  However, note that this equation should not be interpreted as an update rule since $x_t$ appears on both side of the equation.  In fact, this contrast highlights an important difference between OGD and \ourack.

\paragraph{Mahalanobis distance square:} Consider $\Phi(x) = \frac{1}{2} \norm{x}_Q^2$ for positive definite definite $Q$, which is 1-strongly convex and 1-Lipschitz smooth in the Mahalanobis distance $\norm{\cdot}_Q$.  Note that $\nabla \Phi(x) = Qx$.  Then, the first order condition \eqref{eqn: mirror-descent-step} is 
\begin{equation}
x_t = x_{t-1} - \eta_t Q^{-1} \nabla f_t(x_t).
\label{eqn: one-step-ahead-weighted-GD}
\end{equation}
This is analogous to a ``one step ahead" OGD where the underlying metric is a weighted $\ell_2$ metric.

\paragraph{Negative entropy:} If the feasible set is the $\delta$-interior of the simplex $\mathcal{X} = P_\delta = \{ x \mid \sum_{i=1}^n x_i = 1, x_i \ge \delta\}$, and the norm is the $\ell_1$ norm $\norm{\cdot}$, the mirror map defined by the negative entropy $\Phi(x) = \sum_{i=1}^n x_i \log x_i$ is $\frac{1}{2\ln 2}$-strongly convex (by Pinsker's inequality) and $ \frac{1}{\delta \ln 2}$-Lipschitz smooth (by reverse Pinsker's inequality \citep{sason2015}). In this case, $\nabla \Phi(x) = \log x + 1_d$, where $1_d$ represents the all 1s vector in $\mathbb{R}^d$. Then, the first order condition is 
\begin{equation}
x_t =  x_{t-1}\exp(-\eta_t \nabla f_t(x_t)).
\label{eqn: one-step-ahead-multiplicative-weight}
\end{equation}
This can be viewed as a ``one-step ahead'' version of the multiplicative weights update. Again, this equation should not be interpreted as an update rule since $x_t$ appears on both side of the equation.

\section{A Competitive Algorithm}
\label{sec: alg-cr}
%!TEX root = LookingGlass.tex

In this section, we use the \ourack\ framework to give the first algorithm with a dimension-free, constant competitive ratio for online convex optimization with switching costs in general Euclidean spaces, under mild assumptions on the structure of the cost functions.  Recall that, for the most general case, where no constraints other than convexity are applied to the cost functions, Proposition \ref{prop: lowerbound} shows that the competitive ratio of any online algorithm must be $\Omega(\sqrt{d})$ for $\ell_2$ switching costs, i.e., must grow with the dimension $d$ of the decision space. Our goal in this section is to understand when a dimension-free, constant competitive ratio can be obtained.  Thus, we are naturally led to restrict the type of cost functions we consider. 
 
Our main result in this section is a new online algorithm whose competitive ratio is constant with respect to dimension when the cost functions are \emph{locally polyhedral}, a class that includes the form of cost functions used in many applications of online convex optimization, e.g, tracking problems and penalized estimation problems. Roughly speaking, locally polyhedral functions are those that grow at least linearly as one moves away from the minimizer, at least in a small neighborhood. 

\begin{definition}
A function $f_t$ with minimizer $v_t$ is \textbf{\textit{locally $\alpha$-polyhedral}} with respect to the norm $\| \cdot \|$ if there exists some $\alpha, \epsilon > 0$, such that for all $x \in \mathcal{X}$ such that $\norm{x - v_t} \le \epsilon$, $f_t(x) - f_t(v_t)\ge  \alpha \norm{x - v_t}$.
\label{locallypolyhedraldef}
\end{definition}

Note that all strictly convex functions $f_t$ which are locally $\alpha$-polyhedral automatically satisfy $f_t(x) - f_t(v_t) \ge \alpha \norm{x - v_t}$ for all $x$, not just those $x$ which are $\epsilon$ close to the minimizer $v_t$. In this setting, local polyhedrality  is analogous to strong convexity; instead of requiring that the cost functions grow at least quadratically as one moves away from the minimizer, the definition requires that cost functions grow at least linearly. The following examples illustrate the breadth of this class of functions. One important class of examples are functions of the form $\|x - v_t \|_a$ where $\| \cdot \|_a$ is an arbitrary norm; it follows from the equivalence of norms that such functions are locally polyhedral with respect to any norm. Intuitively, such functions represent ``tracking'' problems, where we seek to get close to the point $v_t$. Another important example  is the class $f(x_t) = g(x_t) + h(x_t)$ where $g$ is locally polyhedral and $h$ is an arbitrary non-negative convex function whose minimizer coincides with that of $g$; since $f(x_t) - f(v_t) \geq g(x_t) - g(v_t)$, $f$ is also locally polyhedral. This lets us handle interesting functions such as $f(x_t) = \|x_t\|_1 + x_t' Q x_t$ with $Q$ psd, or even $f(X_t) = 2\|X_t\|_{\infty}  -\log{\det{\left( I + X_t \right)}}$ where the decision variable $X_t$ is a PSD matrix. Note that locally polyhedral function have previously been applied in the networking community, e.g., by \cite{huang2011} to study delay-throughput trade-offs for stochastic network optimization and by \cite{lin2012} to design online algorithm for geographical load balancing in data centers.   
 
Let us now informally describe how the \ouralg\ framework described in Section \ref{sec: alg-meta} can be instantiated to give a competitive online algorithm for locally polyhedral cost functions.  \ouralg\ is, in some sense, \textit{lazy}: instead of moving directly towards the minimizer $v_t$, it moves to the closest point which results in a suitably large decrease in the hitting cost. This can be interpreted as projecting onto a sublevel set of the current cost function. The trick is to make sure that not too much switching cost is incurred in the process. This is accomplished by carefully picking the sublevel set so that the hitting costs and switching costs are balanced. A formal description is given Algorithm \ref{alg: online-projection-CR}. By Lemma \ref{lem: continuity-distance}, step \ref{projection-step-cr} can be computed efficiently via bisection on $l$. Note that the memoryless algorithm proposed in \cite{bansal2015} can be seen as a special case of Algorithm \ref{alg: online-projection-CR} when the decision variables are scalar. %Additionally note that Algorithm \ref{alg: online-projection-CR} can be implemented efficiently using the bisection method presented in Section \ref{sec: alg-meta}.

\begin{algorithm}[t]
	\begin{algorithmic}[1]
% 		\REQUIRE Initial state $x_0$, mirror map $\Phi$, balance parameter $\beta \in (\frac{2\sqrt{\kappa}-1}{\alpha}, 1)$.
			\FOR{$t=1, \ldots, T$}
                  \STATE Observe cost function $f_t$, set $v_t = \argmin_{x} f_t(x)$.
			\IF{$\norm{x_{t-1}-v_t} < \beta f_t(v_t)$}
            \STATE Set $x_t = v_t$ 
            \ELSE
			\STATE Let $x(l) = \Pi_{K_t^l}^\Phi (x_{t-1})$,
            %\argmin_{x \in K_l} D_\Phi(x, x_{t-1})$, 
            increase $l$ until $\norm{x(l) - x_{t-1}} = \beta l$. Here $K_t^l$ is the $l$-sublevel set of $f_t$, i.e., $K_t^l = \{ x \mid f_t(x) \le l\}$. 
			\label{projection-step-cr}
            \STATE $x_t = x(l)$.
            \ENDIF
			\ENDFOR
	\end{algorithmic}
\caption{(Primal) \ouralg}
\label{alg: online-projection-CR}
\end{algorithm}

%\adam{rewrite the discussion of the algorithm so that it makes sense indpendently of the pseudocade.  Basically, write it like "To apply the meta-algorithm in this context we ... This is summarized in Algorithm 2} The key step in Algorithm \ref{alg: online-projection-CR} is step \ref{projection-step-cr}, where we carefully pick a sublevel set so as to precisely balance the hit cost $f_t(x_t)$ and the movement cost $\norm{x_t - x_{t-1}}$; the parameter $\beta$ trades off between the two. 

The main result of this section is a characterization of the competitive ratio of Algorithm \ref{alg: online-projection-CR}.

% The key idea of choosing $K_l$ in step \ref{alg: gradient-step} of Algorithm \ref{alg: online-projection} is to strike balance between movement cost and hit cost (or size of the gradient). In the following lemma we show that there exist an $l$ that achieve such balance, moreover, we find such $l$ efficiently via bisection. The proof is included in the Appendix. 
% %Note that the hit cost $f_t(x_t) = l$, and the movement cost $\norm{x_t - x_{t-1}} = \norm{x_{t-1}, K_l}$ where we define the distance from a point to a convex set $d(u, K) = \min_{v \in K} \norm{u-v}$. 
% \begin{lemma}
% 		Let $x(l) = \argmin_{K_l} D_{\Phi}(x, x_{t-1}) $. The function $g(l) = \norm{x(l) - x_{t-1}}$ is continuous in $l$. 
% 		\label{lem: continuity-distance}
% \end{lemma}

% Given Lemma \ref{lem: continuity-distance}, let $l_0 = f_t(v_t)$, and $l_1 = f_t(x_{t-1})$,  we know that $ g(l_0) > \beta l_0 $, and $0 = g(l_1) < \beta l_1$. By intermediate value theorem, there exists $l_0 < l < l_1$ such that $ g(l) = \beta l$, i.e., $\norm{x_t - x_{t-1}} = \beta f_t(x_t)$.

% Further, by continuity of the function  $ g(l) - \beta l$, we can use bisection to find $l$ such that $g(l) = \beta l$. Computing the projection $\Pi_{K_l}(x_{t-1})$ for each individual $l$ is a convex problem, hence we can compute step \ref{projection-step-cr} in polynomial time independent of $T$.

\begin{theorem}
\label{threeCR}
For every $\alpha > 0$, there exists a choice of $\beta$ such that Algorithm \ref{alg: online-projection-CR} has competitive ratio at most $3 + O(1/\alpha)$ when run on locally $\alpha$-polyhedral cost functions with $\ell_2$ switching costs. More generally, let $\| \cdot \|$ be an arbitrary norm. There exists a choice of $\beta$ such that Algorithm \ref{alg: online-projection-CR} has competitive ratio at most $\frac{\max\{k_2, 1\}}{\min\{k_1, 1\}} \left(3 + O(1/\alpha)\right)$ when run on locally $\alpha$-polyhedral cost functions with switching cost $\| \cdot \|$. Here  $k_1$ and $k_2$ are constants such that $k_1\norm{x} \le \norm{x}_2 \le k_2\norm{x}$.  
\end{theorem}

We note that in the $\ell_2$ setting Theorem \ref{threeCR} has a form which is connected to the best known lower bound on the competitive ratio of memoryless algorithms. In particular, \cite{bansal2015} use a 1-dimensional example with locally polyhedral cost functions to prove the following bound. 

\begin{proposition} No memoryless algorithm can have a competitive ratio less than 3. 
\label{thm: lower-bound}
\end{proposition}

%\noindent Thus, for the $L_2$ setting Algorithm \ref{alg: online-projection-CR} is only $O(1/\alpha)$ from this lower bound. 

Beyond the $\ell_2$ setting, the competitive ratio in Theorem \ref{threeCR} is no longer dimension-free.  It is interesting to note that, when the switching costs are $\ell_1$ or $\ell_{\infty}$ and $\alpha$ is fixed, \ouralg\ has a competitive ratio that is $O(\sqrt{d})$. In particular, we showed in Section \ref{sec: model} that for general cost functions, there is a $\Omega(d)$ lower bound on the competitive ratio for SOCO with $\ell_{\infty}$ switching costs; hence our $O(\sqrt{d})$ result highlights that local polyhedrality is useful beyond the $\ell_2$ case.

%, matching the lower bound in \cite{Friedman1993} and \cite{antoniadis2016}.  \niangjun{Not quite true. That bound is only valid for $l_2$ case. If you go through their example, they showed a competitive ratio of $d / \norm{1_d}$ where $1_d$ is the all 1 vector. So CR is $\sqrt{d}$ if the norm is $l_2$, $d$ if norm is $l_\infty$, and 1 if norm is in $l_1$.}

While Theorem \ref{threeCR} suggests that \ouralg\ has a constant (dimension-free) competitive ratio only in the $\ell_2$ setting, a more detailed analysis shows that it can be constant-competitive outside of the $\ell_2$ setting as well, though for a more restrictive class of locally polyhedral functions.  This is summarized in the the following theorem.

\begin{theorem}\label{generalconstant}
Let $\| \cdot \|$ be any norm such that the corresponding mirror map $\Phi$ has Bregman divergence $D_{\Phi}$ satisfying $\frac{m}{2}\norm{x-y}^2 \le D_{\Phi}(x,y) \le \frac{M}{2}\norm{x-y}^2$  
for all $x, y \in \mathbb{R}^d$ and some positive constants $m$ and $M$. Let $\kappa = M/m$. For every $\alpha > 2\sqrt{\kappa - 1}$, there exists a choice of $\beta$ so that Algorithm \ref{alg: online-projection-CR} has competitive ratio at most $3 + O(1/\alpha)$ when run on locally $\alpha$-polyhedral cost functions with switching costs given by $\| \cdot \|$.
\end{theorem}

This theorem highlights that for any given locally polyhedral cost functions, the task of finding a constant-competitive algorithm can be reduced to finding an appropriate $\Phi$.  In particular, given a class of polyhedral cost functions with $\alpha >0$ and norm $\norm{\cdot}$, the problem of finding a dimension-free competitive algorithm can be reduced to finding a convex function $\Phi$ that satisfies the differential inequality $\frac{1}{2}\norm{x-y}^2\le \Phi(x) - \Phi(y) - \langle \nabla \Phi(y), x - y\rangle \le \frac{\alpha^2+4}{8}\norm{x-y}^2$ for all $x,y \in \mathcal{X}$.

%Now, let us return to the proof of Theorem \ref{threeCR}. Before giving the formal argument, 
We present an intuitive overview of our proof techniques here, and defer the details to the appendix. We use a potential function argument to bound the difference in costs paid by our algorithm and the offline optimal.  Our potential function tracks the distance between the points $x_t$ picked by our algorithm and the points $x_t^*$ picked by the offline optimal. 
%Using a simple triangle inequality argument, we observe we may assume that in each round the offline algorithm has already moved; this simplifies the analysis since we do not need to consider the offline movement cost.
There are two cases to consider. Either the online point or the offline point has smaller hit cost. The first case is easy to deal with, since our algorithm is designed so that the movement cost is at most a constant $\beta$ times the hit cost; hence if our online hit cost is less than the offline algorithm's hit cost, our total per-step cost will at most be a constant times what the offline paid. The second case is more difficult. The key step is Lemma  \ref{lem: potential-change}, where we show that the potential must have decreased if the offline has smaller hit cost. We use this fact to argue that the total per-step cost we charge \ouralg, namely the sum of the hit cost, movement cost, and change in potential, must be non-positive.

The proof of Theorem \ref{generalconstant} parallels that of Theorem \ref{threeCR}. The key difference is the use of a more general form of Lemma \ref{lem: potential-change}, which uses Bregman projection to show that the potential decreases. The Bregman divergence is with respect to the mirror map induced by $\| \cdot \|$. %The rest of the proof is unchanged. The general form of Lemma \ref{lem: potential-change} that is used is stated and proven in the appendix.

\section{A No-regret Algorithm} 
\label{sec: alg-regret}

\begin{algorithm} [t]
\begin{algorithmic}[1]
\FOR{$t=1, \ldots, T$}
			\STATE Define $x(l) = \Pi^\Phi_{K_l}(x_{t-1})$, increase $l$ from $l=f_t(v_t)$, until $\norm{\nabla \Phi(x(l)) - \nabla \Phi(x_{t-1})}_* = \eta \norm{\nabla f_t (x(l))}_*$. 
			\label{projection-step-regret}
			\STATE $x_t =  x(l)$.
			\ENDFOR
\end{algorithmic}
\caption{(Dual) \ouralg}
\label{alg: online-projection-regret}
\end{algorithm}

%We now move from competitive ratio to regret, specifically dynamic regret.  
While the online algorithms community typically focuses on competitive ratio, regret is typically the focus of the online learning community. The difference in performance metrics leads to differences in the settings considered.  In the previous section, we studied locally polyhedral cost functions, while here we focus on cost functions that are continuously differentiable and have a minimizer $v_t$ in the interior of the feasible set $\mathcal{X}$.\footnote{Any convex function can be approximated by a convex functions with these properties, e.g., see \cite{nesterov2005}.% and \cite{wright1999}.
}  

Interestingly, it is has been shown that the change in metric from competitive ratio to regret has a fundamental impact on the type of algorithms that perform well. Concretely, it has been shown that no single algorithm can perform well across (static) regret and competitive ratio \citep{andrew2013}.  Consequently, it is not surprising that we find a different choice of balance in \ourack\ is needed to obtain the no-regret performance guarantees.  Specifically, in contrast to the results of the previous section, which focus on a form of balance in the primal setting, in this section we focus on balance in the dual setting, where we compare costs as measured in the dual norm, $\norm{x}_* = \max_{\norm{z} \le 1} \langle z, x \rangle$. 

We show that choosing  $l$ to balance between the switching cost in the dual space and the size of the gradient leads to an online algorithm with small dynamic regret. It is worth emphasizing that, in contrast to the results of the previous section, we balance the switching cost against the marginal hitting cost $\norm{\nabla f_t(x_t)}_*$ instead of $f_t(x_t)$. A formal description of the instantiation of \ourack\ for regret is given in Algorithm \ref{alg: online-projection-regret}, which can be implemented efficiently via bisection (Lemma \ref{lem: continuity-ratio}).% implies that the right balance can be computed efficiently via bisection when implementing Step \ref{projection-step-regret}. Our main result in this section bounds the dynamic regret of Algorithm \ref{alg: online-projection-regret}. 

\begin{theorem}
Consider $\Phi$ that is an $m$-strongly convex function in $\norm{\cdot}$ with $\norm{\nabla \Phi(x)}_*$ bounded above by $G$ and $\nabla \Phi(0) = 0$. Then the $L$-constrained dynamic regret of Algorithm \ref{alg: online-projection-regret} is $\leq \frac{GL}{\eta}  + \frac{T\eta}{2m}.$
\label{thm: online-projection-regret}
\end{theorem} 

While the result above does not depend on knowing the parameters of the instance, if we know the parameters $T$, $D$ and $L$ ahead of time then we can optimize the balance parameter $\eta$ as follows.

\begin{corollary}\label{cor: online-projection-regret}
When $\eta =\sqrt{\frac{2GLm}{T}}$,  Algorithm \ref{alg: online-projection-regret} has $L$-constrained dynamic regret $\leq \sqrt{\frac{2GLT}{m}}.$ 
\end{corollary}

One interesting aspect of this result is that it has a form similar to the dynamic regret bound on OGD in Theorem 2 of \cite{zinkevich2003}.  Both are independent of the dimension of the decision space $d$, assuming the diameter of the space is normalized to $1$. The key difference is that the bound in Corollary \ref{cor: online-projection-regret} is \emph{independent of the size of the gradients of the cost functions}, unlike in the case of OGD. This can be viewed as a significant benefit that results from the fact that \ourack\ steps in a direction normal to where it lands, rather than where it starts.  

Finally, note that Theorem \ref{thm: online-projection-regret} and Corollary \ref{cor: online-projection-regret} additionally provide bounds on the static regret of \ourack, by setting $L=D$. In this case, Corollary \ref{cor: online-projection-regret} gives a bound of $O(\sqrt{T})$ which matches the lower bound  in the setting where there are no switching costs \citep{hazan2016}.

\bibliography{CR_reference}

\begin{appendix}
%!TEX root = LookingGlass.tex

\section{Lower bounds on competitive ratio and regret}

 To provide insight as to the difficulty of SOCO, here we give an explicit example that yields lower bounds on the competitive ratio an regret achievable in SOCO. Our example is based on the lower bound example for Convex Body Chasing in \cite{Friedman1993}; it was observed in \cite{antoniadis2016} that Convex Body Chasing is a special case of SOCO. Starting from the origin, in each round, the adversary examines the $i$-th coordinate of the algorithm's current action. If this coordinate is negative, the adversary picks the cost function which is the indicator of the hyperplane $x_i = 1$; similarly, if the coordinate is non-negative, the adversary picks the indicator corresponding to the hyperplane $x_i = -1$. Hence our online algorithm is forced to move by at least 1 unit each step, paying a total of at least $d$ cost. The offline optimal, on the other hand, simply moves to the intersection of all these hyperplanes in round 1, paying switching cost $\| (1, 1, \ldots 1) \|$, which is $\sqrt{d}$ when the underlying norm is $\ell_2$ and $1$ when the norm is $\ell_{\infty}$. Hence the competitive ratio is at least $\sqrt{d}$ in the $\ell_2$ setting and $d$ in the $\ell_{\infty}$ setting. Further, the regret is at least $\Omega(d)$ for both the $\ell_2$ and $\ell_{\infty}$ settings. 
Note that while it may appear at first glance that our example requires an adaptive adversary, the same example applies for an oblivious adversary because the online algorithm may be assumed to be deterministic.\footnote{\cite{bansal2015} proves that randomization provides no benefit for SOCO.}

\section{Proof of Lemma \ref{lem: continuity-distance}}
\label{sec: proofs}
Recall the following Pythagorean inequality satisfied by projection onto convex sets using Bregman divergence as measure of distance given by \citet[Lemma 4.1]{bubeck2015}.
\begin{proposition} 
\label{prop: convex-proj-ineq}
For $x_t, x_{t-1}, K_l$ in step \ref{alg: gradient-step} of Algorithm \ref{alg: online-projection}, for any $y \in K_l$, we have 
\[D_{\Phi} (x_t , x_{t-1} ) + D_\Phi ( y, x_t )\le D_\Phi(y, x_{t-1}).\]
\end{proposition} 
In particular, note that when $\Phi$ is the 2-norm squared $\norm{\cdot}_2^2$, $x_t$, $x_{t-1}$ and $y$ form an obtuse triangle. 

	Let $h(l) = D_{\Phi}(x(l), x_{t-1})$, we first show that $h(l)$ is convex, hence continuous in $l$, and from there we show that $g(l)$ is continuous in $l$. To begin, we can equivalently write the variational form 
	\[ h(l)= \min_x \max_{\lambda\ge 0} D_\Phi(x, x_{t-1}) + \lambda(f(x) - l).\]
	Let $H(x, \lambda, l ) = D_\Phi(x, x_{t-1}) + \lambda(f(x) - l)$, $H$ is affine hence convex in $l$ for any given $x$ and $\lambda$. Since maximization preserves convexity, and because $H$ is jointly convex in $(x, l)$, minimization over $x$ also preserves convexity in $l$, hence $h(l) = \min_x \max_{\lambda \ge 0} H(x, \lambda, l)$ is convex hence continuous in $l$. 
	To show $g(l)$ is continuous at $l$, given any $\epsilon >0$, let $\delta>0$ such that $|h(l) - h(l+\delta) | < \epsilon^2 m / 2$ (this can be done as $h$ is continuous in $l$), then 
	\begin{align*}
	 | g(l) - g(l+\delta) | & \le \norm{x(l) - x(l+\delta)} \\
	 & \le \sqrt{ \frac{2}{m} \cdot D_\Phi(x(l+\delta), x(l))} \\
	 & \le \sqrt{ \frac{2}{m} \cdot (D_{\Phi}(x(l), x_{t-1}) - D_\Phi(x(l+\delta), x_{t-1})) } \\
	 & = \sqrt{ \frac{2}{m} \cdot | h(l) - h(l+\delta)| } < \epsilon,
	 \end{align*}
	 where the first inequality is due to triangle inequality, and the second inequality is due to the definition of Bregman divergence and $\Phi$ being $m$ strongly convex in $\norm{\cdot}$, the third inequality is due to the fact that $x(l) \in K_{l+\delta}$ and Proposition \ref{prop: convex-proj-ineq}. Therefore $g$ is a continuous function in $l$.

\section{Proof of Lemma \ref{lem: continuity-ratio}}
First, we will show that $h_1(l) = \norm{\nabla \Phi(x_{t-1})- \nabla\Phi(x(l))}_*$ is continuous in $l$. For any $l_1, l_2$, $ |h_1(l_1) - h_1(l_2)| \le \norm{\nabla \Phi(x(l_1)) - \nabla \Phi(x(l_2)) }_*$ by the triangle inequality, as $\Phi$ is continuously differentiable, we only need to show that for any $l$, given any $\epsilon>0$, we can find $\delta>0$, such that for all $l'$, such that $|l - l'|<\delta$, $\norm{\nabla \Phi (x(l)) - \nabla \Phi (x(l'))}_* <\epsilon$. By Lipschitz smoothness of $\Phi$, we have:
\begin{equation}
D_\Phi(x, y) \ge \frac{1}{2M} \norm{\nabla \Phi(x) - \nabla \Phi(y) }_*^2. 
\label{eqn: smoothness-ineq}
\end{equation}

To show \eqref{eqn: smoothness-ineq}, note that by \cite[Theorem 1]{kakade2009}
 $\Phi$ is $M$-Lipschitz smooth w.r.t $\norm{\cdot}$ if and only if $\Phi^*$ is $\frac{1}{M}$-smooth w.r.t $\norm{\cdot}_*$. Let $u = \nabla \Phi(x), v = \nabla \Phi(y)$, then by strong convexity, 
 \begin{equation}
  \Phi^*(v) - \Phi^*(u) - \langle \nabla \Phi^*(u), v - u \rangle \ge \frac{1}{2M} \norm{ v - u}_*^2.
  \label{eqn: strong-convex-ineq}
  \end{equation}
 By the Fenchel inequality and the definition of $u, v$, we have $\Phi^*(v) = \langle v, y \rangle - \Phi(y)$ and $\Phi^*(u) = \langle u, x \rangle - \Phi(x)$. Furthermore, $\nabla \Phi^*(u) = \nabla \Phi^* (\nabla \Phi(x)) = x$, substituting these into \eqref{eqn: strong-convex-ineq} gives \eqref{eqn: smoothness-ineq}.
  
 % \niangjun{The Euclidean norm case of \eqref{eqn: smoothness-ineq} can be found in \cite[Lemma 3.5]{bubeck2015}, I have yet to find a source I can cite for the non-Euclidean case. A potential problem here is that \cite{kakade2009} is unpublished, which may be problematic, but the authors are quite established though. I wonder if there is a sleeker proof out there...}

Using \eqref{eqn: smoothness-ineq} and Proposition \ref{prop: convex-proj-ineq}, we can upper bound $ \norm{\nabla \Phi(x(l) - \nabla \Phi(x(l')}_*$ by 
\begin{align*}
	\norm{\nabla \Phi(x(l)) - \nabla \Phi(x(l'))}_* &\le \sqrt{2M D_\Phi(x(l), x(l'))} \\
	&\le \sqrt{2M }\sqrt{ D_\Phi(x(l), x_{t-1} ) - D_\Phi(x(l'), x_{t-1}) }.
\end{align*}
However, since we have already shown that $D_\Phi(x(l), x_{t-1})$ is continuous in $l$ in the proof of Lemma \ref{lem: continuity-distance}, we can choose $l'$ sufficiently close to $l$ to make the right hand side smaller than $\epsilon$. Therefore $h_1(l) = \norm{\nabla \Phi(x_{t-1})- \nabla\Phi(x(l))}_*$ is continuous in $l$. Second, since $f_t$ is also continuously differentiable, and by the continuity of $x(l)$ in $\norm{\cdot}$, $h_2(l) = \norm{\nabla f_t(x(l))}_*$ is also continuous in $l$. Thus, the ratio $h(l) = \frac{h_1(l)}{h_2(l)}$ is continuous in $l$.

\section{Proof of Theorem \ref{threeCR}}
%\begin{proof}{\textbf{of Theorem \ref{threeCR}.}}
We first consider the case when the switching cost is the $\ell_2$ norm. We define $H_t = f_t(x_t), M_t =\norm{x_t - x_{t-1}}$, and define $H_t^*$ and $M_t^*$ analogously. We use the potential function $\phi(x_t, x_t^*) = C \norm{x_t - x_t^*}$; $C$ will end up being the competitive ratio of Algorithm \ref{alg: online-projection-CR}. To show Algorithm \ref{alg: online-projection-CR} is $C$-competitive, we need to show that for all $t$, 
\begin{equation}
 	H_t + M_t + \phi(x_t, x_t^*) - \phi(x_{t-1}, x_{t-1}^*) \le C(H_t^* + M_t^*),
    \label{eqn: potential-ineq}
\end{equation}
then summing up the inequality over $t$ implies the result. 
To begin, applying the triangle inequality, we see  

\begin{align}
	\phi(x_t, x_t^*) - \phi(x_{t-1}, x_{t-1}^*) &\le C  (\norm{x_t^* - x_t} - \norm{x_t^* - x_{t-1}}) + C M_t^*.
    \label{eqn: potential-change}
\end{align}

% \begin{align}
% 	\phi(x_t, x_t^*) - \phi(x_{t-1}, x_{t-1}^*) &= C (\norm{x_t - x_t^*} - \norm{x_{t-1} - x_{t-1}^*}) \notag \\
% 	&= C ( \norm{x_t - x_t^*} - \norm{x_t^* - x_{t-1}} + \norm{x_t^* - x_{t-1}} -  \norm{x_{t-1} - x_{t-1}^*}) \notag \\
% 	&\le C  (\norm{x_t - x_t^*} - \norm{x_t^* - x_{t-1}}) + C M_t^*
%     \label{eqn: potential-change}
% \end{align}

\noindent Combining \eqref{eqn: potential-change} and \eqref{eqn: potential-ineq}, we see that, to show Algorithm \ref{alg: online-projection-CR} is $C$-competitive, it suffices to show 
\begin{align}
	H_t + M_t + C (\norm{x_t - x_t^*} - \norm{x_t^* - x_{t-1}}) \le C H_t^*, \label{eqn: proofgoal}
\end{align}
Notice that we always have $M_t \leq \beta H_t$.  We divide our analysis into the following two cases:
\begin{enumerate}
	\item $H_t \le H_t^*$: Since $M_t \le \beta H_t$, by the triangle inequality we have
	\begin{align*} 
		&H_t + M_t + C (\norm{x_t^* - x_t} - \norm{x_t^* - x_{t-1}}) \\
		\le & H_t + (1+C) M_t \le (1 + \beta(1+C))H_t \le CH_t \le CH_t^*,
	\end{align*}
	where in the penultimate step we assumed that $\beta$ was picked so that $ 1+\beta( 1 + C) \le C$. 
	\item $H_t > H_t^*$: In this case, it must true that $M_t = \beta H_t$, since $H_t^*$ is strictly smaller than $H_t$, implying that our algorithm did not reach the minimizer $v_t$.  We use the following Lemma, proved in the Appendix, which shows that the change in potential must actually be negative in this case.
% \begin{lemma}
% 	For Algorithm \ref{alg: online-projection-CR}, when $H_t > H_t^*$ and $f_t(x) \ge \alpha\norm{x - v_t}$, we have 
% 	\[ \norm{x_t - x_t^*} - \norm{x_t^* - x_{t-1} } \le -\gamma \norm{x_t - x_{t-1}}, \]
% 	where $\gamma = \frac{1}{\sqrt{\kappa}}\sqrt{1 + \left( \frac{2}{\alpha\beta}\right)^2} - \frac{2}{\alpha\beta}$, and $\gamma>0$ when $\beta > \frac{2\sqrt{\kappa - 1}}{\alpha}$.
% 	\label{lem: potential-change}
% \end{lemma}

\begin{lemma}
	For Algorithm \ref{alg: online-projection-CR}, when $H_t > H_t^*$ and $f_t(x) \ge \alpha\norm{x - v_t}$, we have 
	\[ \norm{x_t - x_t^*} - \norm{x_t^* - x_{t-1} } \le -\gamma \norm{x_t - x_{t-1}}, \]
	where $\gamma = \sqrt{1 + \left( \frac{2}{\alpha\beta}\right)^2} - \frac{2}{\alpha\beta}$.
	\label{lem: potential-change}
\end{lemma}

\noindent Using Lemma \ref{lem: potential-change}, we have 
\begin{align*} 
	&H_t + M_t + C (\norm{x_t^* - x_t} - \norm{x_t^* - x_{t-1}}) 
	\le & H_t + M_t - C \gamma M_t = (1 + \beta(1 - C \gamma)) H_t
\end{align*}
To show \eqref{eqn: proofgoal}, it suffices to pick $\beta$ such that $ 1 + \beta \left (1 - C \gamma \right) \le 0$.
%which is equivalent to $C \ge \frac{1+\beta}{\beta} \cdot \frac{1}{\gamma}$, and $\gamma = \frac{1}{\sqrt{\kappa}}\sqrt{ 1 + \left(\frac{2}{\alpha\beta}\right)^2} - \frac{2}{\alpha\beta}$ from Lemma \ref{lem: potential-change}, 
%$(\dagger)$ holds in this case if 
%\[
%C \ge \frac{1+\beta}{\beta} \left(\sqrt{\left(\frac{2}{\alpha\beta}\right)^2+1}+\frac{2}{\alpha\beta}\right),
%\]
\end{enumerate}
Combining the inequalities obtained in cases 1 and 2, we conclude that for any $\beta \in (0, 1)$, Algorithm \ref{alg: online-projection-CR} is $C$-competitive, where $$C = \max{ \left( \frac{1 +\beta}{1 - \beta}, \frac{1 +\beta}{\beta}\frac{1}{\gamma} \right). }$$ Note that the first term is increasing in $\beta$ and tends to $+\infty$ as $\beta \rightarrow 1$, and the second is decreasing in $\beta$ and tends to $+\infty$ as $\beta \rightarrow 0_+$; hence to minimize $C$ we should pick $\beta$ so that both terms are equal. We easily obtain $\beta = \frac{1}{2} + \frac{1}{\alpha + 2}$, and $C = 3 + 8/\alpha$.

%increasing in $\eta$ and use the constraint to obtain $\eta = \frac{1 + \beta}{\beta}\frac{1}{\gamma}$. The minimization problem thus becomes $\min_{\beta \geq 0} \max{\left( \frac{1 + \beta}{\beta}\frac{1}{\gamma}, (1 + \beta)(1 + \frac{1}{\gamma}) \right)} $. We observe that the first term is decreasing in $\beta$ and the second is increasing \niangjun{second one is not necessarily increasing, need to revisit}; hence the minimum is attained when the two terms are equal. Solving for $\beta$, we obtain $\beta = \frac{1}{2} + \frac{1}{\alpha + 2}$ which immediately gives  $C = 3 + 8/\alpha$.

This result easily extends to the case when the switching cost is an arbitrary norm $\| \cdot \|_a$. Since in finite dimensions all norms are equivalent, we know that there exist some $k_1, k_2 > 0$ such that $k_1\norm{x} \le \norm{x}_2 \le k_2\norm{x}$ . We immediately obtain
\begin{align}
\label{eqn: norm1}
\sumt f_t(x_t) + \norm{x_t - x_{t-1}} &\ge \min\{1, k_1\} \left(\sumt f_t(x_t) + \norm{x_t - x_{t-1}}_a\right) \\
\label{eqn: norm2}
\sumt f_t(x_t^*) + \norm{x_t^* - x_{t-1}^*} &\le \max\{1, k_2\} \left(\sumt f_t(x_t^*) + \norm{x_t^* - x_{t-1}^*}_a\right) 
\end{align}
Combining \eqref{eqn: norm1} and \eqref{eqn: norm2} with the previous observation that the online cost is at most $C$ times the offline cost finishes the proof.
%\end{proof}

\section{Proof of Lemma \ref{lem: potential-change}}
In this section, we actually prove a more general statement, from which Lemma \ref{lem: potential-change} follows.  This more general statement is not needed to prove Theorem \ref{threeCR}, but is needed in our proof of Theorem \ref{generalconstant}. 

\begin{lemma} \label{lem: general}
Let $\| \cdot \|$ be any norm such that the corresponding mirror map $\Phi$ has Bregman divergence satisfying $\frac{m}{2}\norm{x-y}^2 \le D_{\Phi}(x,y) \le \frac{M}{2}\norm{x-y}^2.$
Let $\kappa = M/m$. For Algorithm \ref{alg: online-projection-CR}, when $H_t > H_t^*$ and $f_t(x) \ge \alpha\norm{x - v_t}$, we have $ \norm{x_t - x_t^*} - \norm{x_t^* - x_{t-1} } \le -\gamma \norm{x_t - x_{t-1}},$	where $\gamma = \frac{1}{\sqrt{\kappa}}\sqrt{1 + \left( \frac{2}{\alpha\beta}\right)^2} - \frac{2}{\alpha\beta}$.
\end{lemma}

\noindent Before turning to the proof, we note that $\kappa = 1$ when the norm is $\ell_2$, recovering Lemma \ref{lem: potential-change} used in the proof of Theorem \ref{threeCR}. 

\begin{proof}{\textbf{of Lemma \ref{lem: general}.}} By the triangle inequality, $\norm{x_t - x_t^*} \le \norm{x_t - v_t} + \norm{x_t^* - v_t}$. To upper bound $\norm{x_t - x_t^*},$ we separately bound each term on the right hand side.  By the assumption that $f_t$ is polyhedral, $\norm{x_t - v_t} \le \frac{1}{\alpha} f_t(x_t) = \frac{1}{\alpha}H_t$. Also, when $H_t^* < H_t$, we must have $M_t = \beta H_t$, hence $\norm{x_t - v_t} \le \frac{1}{\alpha\beta} M_t.$ Using similar argument together with the fact that $H_t^* < H_t,$ we have $\norm{x_t^* - v_t} \le \frac{1}{\alpha\beta} M_t$. 
Hence 
\begin{equation}
	\norm{x_t - x_t^*} \le \norm{x_t - v_t} + \norm{x_t^* - v_t} \le \frac{2}{\alpha\beta}M_t.
	\label{eqn: dist1}
\end{equation}

\noindent Since $x_t$ is the projection of $x_{t-1}$ onto the sublevel set $\{x \mid f(x) \le H_t\}$, this set must contain $x_t^*$ since by assumption $H_t^* \le H_t$. Therefore by Proposition \ref{prop: convex-proj-ineq} we have 
\[
 D_\Phi( x_t^*, x_t) + D_\Phi(x_t, x_{t-1}) \le D_\Phi(x_t^*, x_{t-1}), 
 \]
 since $\Phi$ is $m$-strongly convex and $M$ strongly smooth in  $\norm{\cdot}$, 
$\frac{m}{2}\norm{x-y}^2 \le D_{\Phi}(x,y) \le \frac{M}{2}\norm{x-y}^2,$ hence
 \begin{align*}
 \norm{x_t - x_t^*}^2 + \norm{x_t - x_{t-1}}^2 \le \kappa \norm{x_t^* - x_{t-1}}^2.
 \end{align*}
 Let $\norm{x_t - x_t^*} = r M_t$; note that by \eqref{eqn: dist1}, $r \le \frac{2}{\alpha\beta}$. We have, 
\begin{equation}
	\norm{x_t^* - x_{t-1}} \ge \frac{1}{\sqrt{\kappa}} \sqrt{ 1 + r^2} M_t.
	\label{eqn: dist2}
\end{equation} 
Hence 
\begin{equation}
\norm{x_t^* - x_{t-1}} -  \norm{x_t - x^*_t} \ge \left (\frac{1}{\sqrt{\kappa}} \sqrt{ 1 + r^2}  - r\right) M_t.
\end{equation} Since $\frac{1}{\sqrt{\kappa}} \le 1$, $\frac{1}{\sqrt{\kappa}}\sqrt{ 1 + r^2}  - r$ is a decreasing function in $r$; this can seen by taking the derivative with respect to $r$. Combining this together with the fact that $r \le \frac{2}{\alpha\beta}$ we have $$\norm{x_t - x^*_t} - \norm{x_t^* - x_{t-1}} \le -\gamma M_t$$ for $\gamma = \frac{1}{\sqrt{\kappa}}\sqrt{ 1 + \left(\frac{2}{\alpha\beta}\right)^2} - \frac{2}{\alpha\beta}$. Note that $\gamma>0$ when $\beta > \frac{2\sqrt{\kappa-1}}{\alpha}$. 
\end{proof}

\section{Proof of Theorem \ref{thm: online-projection-regret}}
%\begin{proof}{\textbf{of Theorem \ref{thm: online-projection-regret}.}} 
Recall that we can write the update rule as: 
\[ \nabla \Phi(x_t) = \nabla \Phi(x_{t-1}) - \eta \nabla f_t(x_t),\] 
Let $\{x_t^L\}_{t=1}^T$ denote a $L$-constrained offline optimal solution. By convexity of $f_t$, we have 
\begin{align} 
f_t(x_t) &- f_t(x_t^L) \le \langle \nabla f_t(x_t), x_t - x_t^L\rangle = \frac{1}{\eta} \langle \nabla \Phi(x_{t-1}) - \nabla \Phi(x_t) , x_t - x_t^L\rangle\notag \\
  = & \frac{1}{\eta}\Big(\langle \nabla \Phi(x_{t-1}) - \nabla \Phi(x_t), x_{t-1} - x_t^L \rangle - \langle \nabla \Phi(x_t) - \nabla\Phi(x_{t-1}), x_t - x_{t-1} \rangle \Big) 
  \label{eqn: expansion}
\end{align}
Recall that the Bregman divergence satisfies the equality
\[
\langle \nabla f(x) - \nabla f(y), x-z\rangle = D_f(x,y) + D_f(z, x) - D_f(z, y),
\]
for all $x, y, z \in \mathbb{R}^d$. We use this identity in each of the inner products in \eqref{eqn: expansion} to obtain
\begin{align*}
f_t(x_t) - f_t(x_t^L) 
\leq & \frac{1}{\eta} \Big(  D_\Phi(x_{t-1}, x_t) + D_\Phi(x_t^L, x_{t-1}) \quad - D_\Phi(x_t^L, x_t)  \\
& - D_\Phi(x_t, x_{t-1}) - D_\Phi(x_{t-1}, x_t) + D_\Phi(x_{t-1}, x_{t-1})     \Big) \\
= & \frac{1}{\eta} \Big( D_\Phi(x_t^L, x_{t-1}) - D_\Phi(x_t^L, x_t) \Big) - \frac{1}{\eta} (D_\Phi(x_t, x_{t-1})).
\end{align*}
Notice that 
\begin{align*}
  & D_\Phi(x_t^L, x_{t-1}) - D_\Phi(x_t^L, x_t) \\
= &\Phi(x_t^L) - \Phi(x_{t-1}) - \langle \nabla \Phi(x_{t-1}), x_t^L - x_{t-1}\rangle -\Phi(x_t^L) + \Phi(x_t) +  \langle \nabla\Phi(x_t) , x_t^L-x_t \rangle \\
= & (\Phi(x_t) - \langle \nabla \Phi(x_t), x_t \rangle) - (\Phi(x_{t-1}) - \langle \nabla \Phi(x_{t-1}),  x_{t-1} \rangle) + \langle \nabla \Phi(x_t) - \nabla \Phi(x_{t-1}), x_t^L \rangle \\
= & D_\Phi(0, x_{t-1}) - D_\Phi(0, x_t) + \langle \nabla \Phi(x_t) - \nabla \Phi(x_{t-1}), x_t^L \rangle
\end{align*}
Summing over $t$, we have 
\begin{align*}
& \sumt  D_\Phi(x_t^L, x_{t-1}) - D_\Phi(x_t^L, x_t) \\
= &D_{\Phi}(0, x_0) - D_\Phi(0, x_T) + \sum_{t=1}^{T-1} \langle\nabla \Phi(x_t), x_t^L - x_{t+1}^L \rangle - \langle \nabla \Phi(x_0), x_1^L \rangle + \langle \nabla \Phi(x_T), x_T^L\rangle \\
\le & \sumt \norm{ \nabla \Phi(x_t)}_* \norm{ x_t^L - x_{t+1}^L} \le G \sumt \norm{x_t^L - x_{t+1}^L }= GL,
\end{align*}
where in the penultimate inequality we used the fact that $x_0 = 0$, $\nabla \Phi(0) = 0$, and $x_{T+1}^L = x_0^L = 0$. Putting it all together, we obtain 
\begin{align*}
&\cost(\ourack) - \cost(OPT(L)) \\
= &\sumt \left(f_t(x_t) - f_t(x_t^L)\right) + \sumt \left( \norm{x_t - x_{t-1}} - \norm{x_t^L - x_{t-1}^L}\right) \\
\le &\frac{GL}{\eta}  + \sumt \left(\norm{x_t - x_{t-1}} - \frac{1}{\eta} D_\Phi(x_t, x_{t-1}) \right) - L \\
\le &\frac{GL }{\eta}  + \sumt \left(\norm{x_t - x_{t-1}} - \frac{m}{2\eta} \norm{x_t - x_{t-1}}^2 \right) 
\le \frac{ GL }{\eta} + \frac{\eta T}{2m}, 
\end{align*}
where the last inequality is due to completing the square and throwing away the negative terms.

\end{appendix}

\end{document}